\documentclass[conference]{IEEEtran}
\IEEEoverridecommandlockouts

\usepackage{cite}
\usepackage{amsmath,amssymb,amsfonts}
\usepackage{algorithmic}
\usepackage{graphicx}
\usepackage{textcomp}
\usepackage{xcolor}
\usepackage{amsmath,amssymb,booktabs}
\usepackage{graphicx}
\usepackage{url}
\usepackage{cite}

\def\BibTeX{{\rm B\kern-.05em{\sc i\kern-.025em b}\kern-.08em
    T\kern-.1667em\lower.7ex\hbox{E}\kern-.125emX}}
\begin{document}

\title{
EnergyEminence: Source-Aware Environmental Calibration and Evaluation in a Physics-Grounded Grid Digital Twin\\
}

\author{
\IEEEauthorblockN{
Huy Trinh\IEEEauthorrefmark{1}\IEEEauthorrefmark{2},
Michael Mai\IEEEauthorrefmark{2},
Yu Nong\IEEEauthorrefmark{2}
}
\IEEEauthorblockA{
\IEEEauthorrefmark{1}\textit{University of Waterloo}, Waterloo, ON, Canada\\
\IEEEauthorrefmark{2}\textit{Kraftgene AI Inc.}, Toronto, ON, Canada\\
michel.trinh@kraftgeneai.ca,\;
m.mai@kraftgeneai.ca,\;
yu.nong@kraftgeneai.ca
}
}

\maketitle

\begin{abstract}

Power-grid digital twins must combine data-driven prediction with physically meaningful state evolution while preserving the provenance of environmental observations. This paper presents an early-stage EnergyEminence testbed that couples an IEEE 118-bus-style graph-temporal predictor, nonlinear AC cascade simulation, and operator-dashboard alike temporal replay. In addition, we introduce a shared bounded calibration that converts wildfire-detection confidence and spatial extent into source-comparable wildfire interpretable and explainable evidence. We then evaluate it with visually diverse fire and hard-negative videos. Sixteen synthetic environmental videos are curated to generate 160 source-tracked grid scenarios, and a source-video-disjoint test yields 10 true positives, 8 false positives, 22 true negatives, and no false negatives. The errors occur in stressed, non-cascading scenarios conditioned on an unseen growing-fire source. Our diagnostic then reveals environmental shortcut learning that is obscured by scenario-level random splitting. The paper therefore contributes a data-centric and inspectable evaluation workflow for multimodal grid-resilience models, together with evidence supporting separation of environmental alerting from electrical cascade inference.

\end{abstract}


\begin{figure*}[!t] 
  \centering
  \includegraphics[
    width=\textwidth,
    keepaspectratio,          
    trim={10 160 30 160pt},        
    clip
  ]{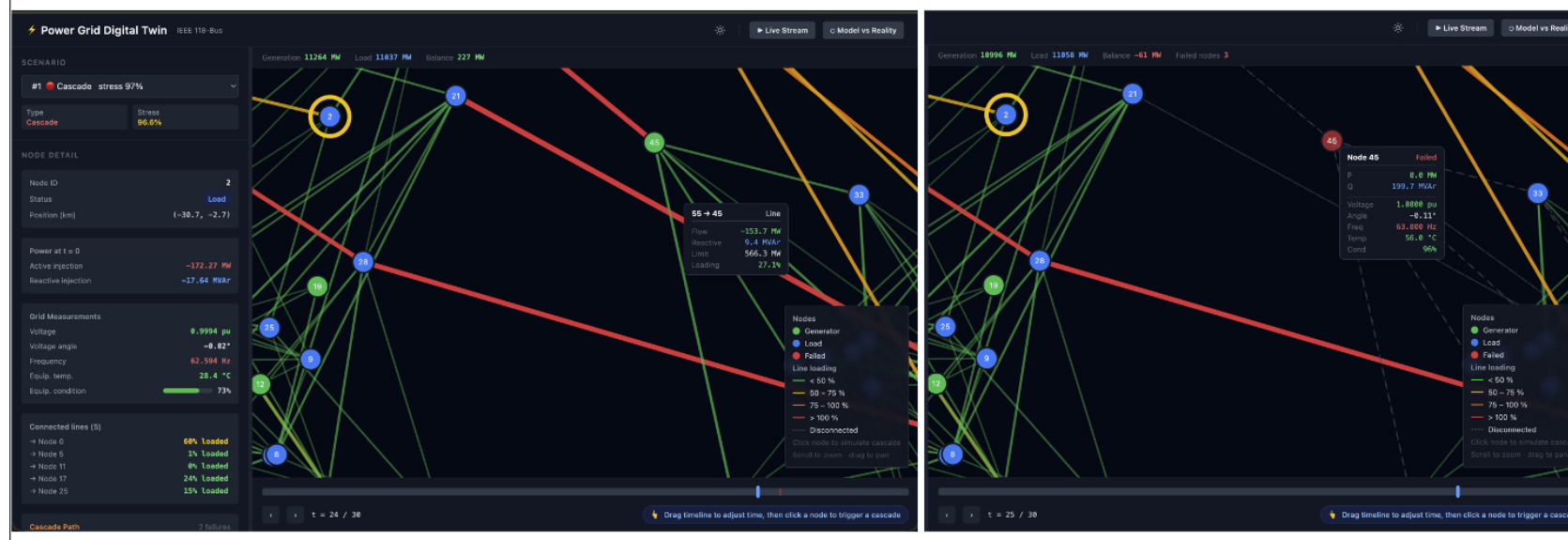}
  
  \caption{EnergyEminence replay for Scenario~\#1 (cascade, displayed stress approximately 97\%). \textbf{Left}: $t=24/30$, immediately before onset, with node telemetry, adjacent line loading, and positive generation--load balance. \textbf{Right}: $t=25/30$, where the displayed balance is negative, three nodes are failed, and affected connections are rendered as failed or disconnected.}
\label{fig:interface}
\end{figure*}
\section{Introduction}
Cascading failures are spatial, temporal and state-dependent events: an initiating outage redistributes power, operating limits may be crossed, and later failures can occur at electrically coupled assets. 
Nonlinear contingency simulation is therefore central to resilience analysis, while graph neural networks (GNNs) provide a complementary surrogate as the grid is naturally represented by buses and assets connected through electrical branches~\cite{10333943}.
Digital twins extend this objective beyond prediction by maintaining a time-indexed representation that links measurements, models, simulation, visualization, and operator decisions. Recent reviews and studies identify monitoring, operational optimization, asset management, and disaster response as major power-grid digital-twin applications \cite{mchirgui2024,SUBRAMANIAN2026109218}. Wildfire-oriented systems similarly combine geospatial sensing, AI detection, and virtual scenario analysis to support earlier intervention \cite{11385902}. However, an environmental observation such as smoke or flame is not equivalent to an electrically unstable state. Additionally, when synthetic grid scenarios are repeatedly generated from a small video corpus, source-specific visual patterns can also appear in multiple data partitions and degrade the generalization. Related video-object-detection research has shown that random allocation of highly correlated frames can create information leakage~\cite{10485397}.  We then investigate these problems in EnergyEminence\footnote{The EnergyEminence demo is available online at \url{https://energyeminence.online/}}, which instantiates a multimodal environmental--infrastructure cascade framework~\cite{doi:10.36227/techrxiv.176288076.69687745/v2} as an interactive physics-grounded testbed. The present paper extends it with source-comparable environmental calibration and source-aware evaluation. 
The paper makes two main contributions:
\begin{itemize}
    \item a source-comparable calibration from video detections to bounded environmental evidence, evaluated with no-fire, fire, and smoke-like hard-negative sources; and
    \item a source-video-disjoint scenario protocol that reveals a growing-fire shortcut failure mode hidden by scenario-level random splitting.
\end{itemize}
\section{EnergyEminence Testbed and Method}
\subsection{Shared State and Dual Inference Paths}
Figure~\ref{fig:interface} shows the EnergyEminence's interface with a scenario browser on the left to select from normal, stressed, or cascading trajectory. At the bottom, the shared timeline exposes system balance, node telemetry, line loading, environmental video, and predicted or simulated failures on one topology.
At timestep $t$, the inspectable digital-twin state is written as
\begin{equation}
\mathcal{S}_t=\bigl(\mathcal{G},\mathbf{X}_V(t),\mathbf{X}_E(t),s_t,\mathcal{F}_t\bigr),
\qquad \mathcal{G}=(\mathcal{V},\mathcal{E}),
\label{eq:state}
\end{equation}
where $\mathcal{V}$ and $\mathcal{E}$ are grid assets and electrical connections, $\mathbf{X}_V(t)$ and $\mathbf{X}_E(t)$ contain node and branch measurements, $s_t$ is environmental evidence, and $\mathcal{F}_t$ is the failed-component set. The node features include active/reactive injection, voltage, frequency, temperature, condition, and failure state. And the branch features include active/reactive flow, loading, limits, and connectivity. The learned and physics paths consume the same state but proceed differently:  
\begin{equation}
(\widehat{\mathbf p},\widehat{\boldsymbol\tau},\widehat{\mathbf r})
=f_{\theta}(\mathcal{S}_{t_0:t_o}),
\qquad
\mathcal{S}_{t+1}=\Phi_{\mathrm{AC}}(\mathcal{S}_t,\mathcal{F}_t).
\label{eq:dual}
\end{equation}
Here, $f_{\theta}$ combines graph attention for electrically connected context with a multi-layer Long Short-Term Memory (LSTM) for temporal evolution and returns node-failure probabilities $\widehat{\mathbf p}$, failure times $\widehat{\boldsymbol\tau}$, and risk factors $\widehat{\mathbf r}$. The simulator $\Phi_{\mathrm{AC}}$ applies an initiating failure, recomputes nonlinear AC power flow, and advances the failed set when configured loading, voltage, frequency, or thermal limits are violated. Figure~\ref{fig:interface} deliberately shows one transition immediately before and after cascade onset: the fixed topology and common clock make the learned and simulated paths directly interpretable and inspectable.

\subsection{Source-Comparable Environmental Evidence}
\begin{figure*}[!t] 
  \centering
  \includegraphics[
    width=\textwidth,
    keepaspectratio,          
    trim={0 350 0 350pt},        
    clip
  ]{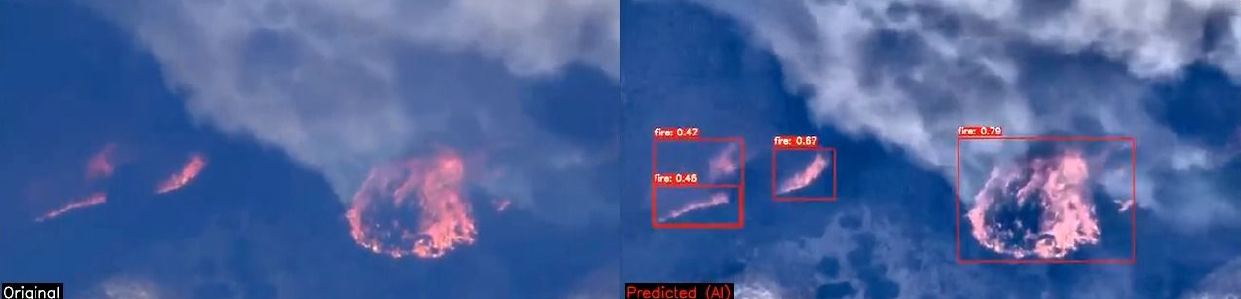}
  
\caption{Qualitative example of frame-level wildfire perception used by the environmental-evidence pipeline. Left: original video frame containing multiple visible fire regions. Right: detections produced by the YOLOv8n wildfire detector, with bounding boxes and confidence scores. The detector outputs retain both confidence and spatial extent before being converted to the shared bounded evidence signal in~\eqref{eq:evidence}. This visualization illustrates the perception input to EnergyEminence; it is not itself an evaluation of cascade-prediction accuracy.} \label{fig:wildfire_detection}
\end{figure*}

Accepted bounding boxes are produced by the framework's YOLOv8n wildfire detector. For detection $b$ in frame $t$, the evidence score is
\begin{equation}
\begin{aligned}
q_{t,b}&=w_{c_b}p_{t,b}
\left(\frac{A_{t,b}}{A_{\mathrm{frame}}}\right)^{\alpha},\\
e_t&=\max_b q_{t,b}+\beta\!\sum_{b\neq b^{\star}}q_{t,b},
\qquad s_t=1-\exp(-\gamma e_t).
\end{aligned}
\label{eq:evidence}
\end{equation}
The detector confidence is $p_{t,b}$, $A_{t,b}/A_{\mathrm{frame}}$ is relative box area, $w_{c_b}$ is a class weight, and $\alpha$ controls area sensitivity. The strongest detection $b^{\star}$ is retained fully, and secondary detections are reduced by $\beta$, and $\gamma$ sets one shared monotonic evidence scale. Therefore, zero evidence remains zero, and each source is not independently forced to become one. The generator spatially attenuates this evidence from sampled fire location $\mathbf p_f$ to node position $\mathbf p_v$:
\begin{equation}
r_{v,t}=s_t\exp\!\left(-\|\mathbf p_v-\mathbf p_f\|_2/20\right).
\label{eq:spatial}
\end{equation}
A representative detector output is shown in Fig.~\ref{fig:wildfire_detection}, where the retained bounding-box areas and confidence scores correspond to the quantities $A_{t,b}$ and $p_{t,b}$ in~\eqref{eq:evidence}.
\subsection{Synthetic Corpus and Source-Aware Split}
\begin{figure}[htb]
\centerline{\includegraphics[width=\linewidth, keepaspectratio,          
    trim={10 80 10 82pt},        
    clip
    ]{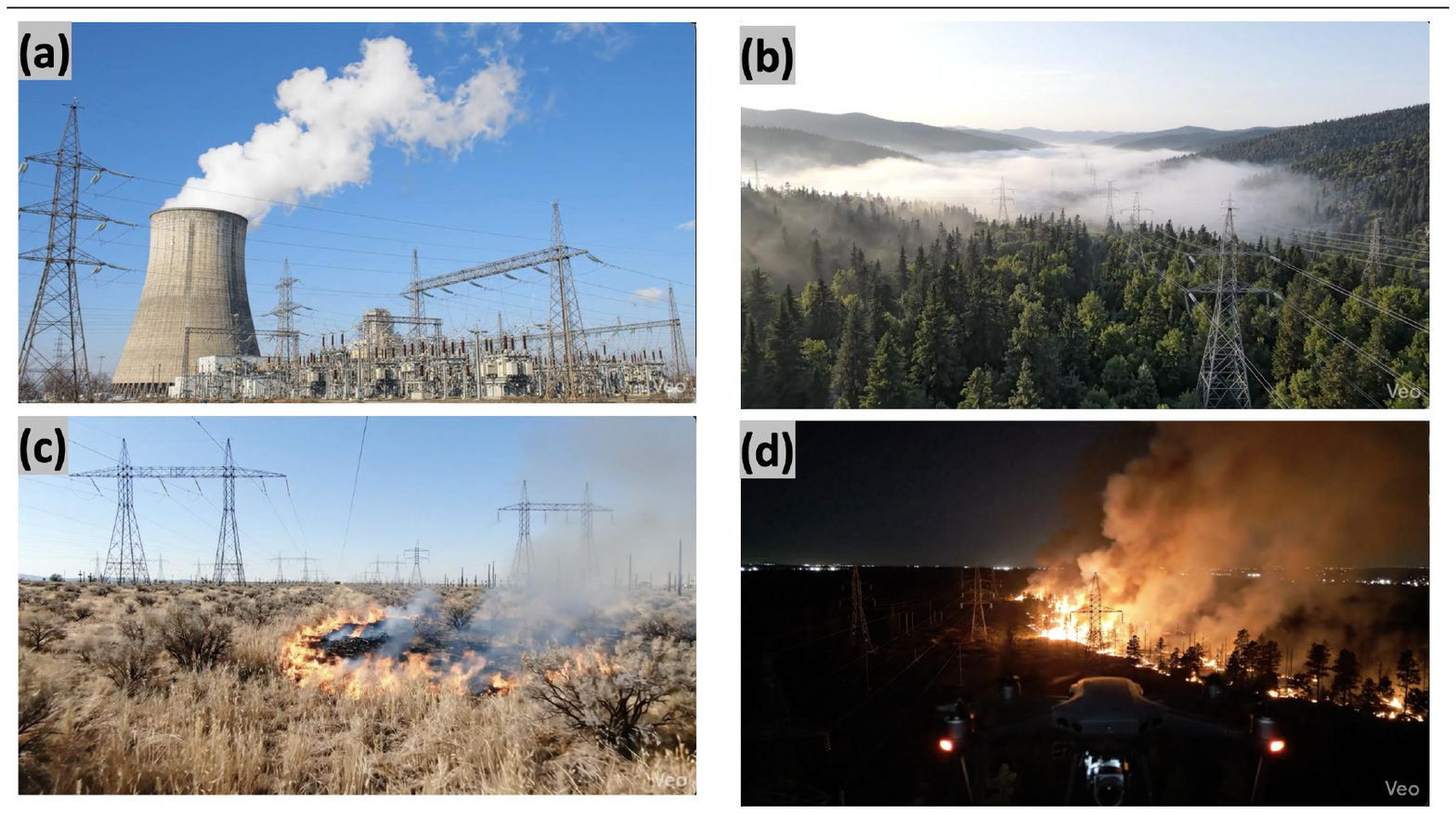}}
\caption{Representative environmental conditions: (a) industrial steam hard negative, (b) fog hard negative, (c) small vegetation fire, and (d) night wildfire. Fog and steam deliberately test smoke-like false alarms; graded fire size and night illumination test source and severity variation.}
\label{fig:videos}
\end{figure}
Figure~\ref{fig:videos} illustrates four representative conditions. Eighteen synthetic clips were generated under six labels: no fire, hard negative, small fire, growing fire, large fire, and night fire. The hard-negative class includes fog, industrial steam, and rain, which can resemble smoke without containing wildfire. Among these, sixteen clips passed signal-quality screening and been used. Ten grid trajectories were generated per accepted source, yielding 160 scenarios with retained source identifiers. We then used the fixed generation gain $\gamma=10$ in the corpus and $\gamma=4$ in three-video mechanism audit in Section~\ref{subsec:calibration_audit} to isolate the normalization effect. The grouped split assigns all scenarios from one video to exactly one partition: 8/4/4 videos, corresponding to 80/40/40 training, validation, and test scenarios. The reported feasibility run used 835,961 trainable parameters, learning rate $10^{-4}$, batch size one, and three CPU epochs. This results in the changes of evaluation question, from recognition of scenarios derived from familiar sources to transfer onto an unseen environmental sequence. The generator intentionally correlates video evidence with grid stress in this feasibility study. Accordingly, the experiment diagnoses shortcut learning in a coupled synthetic pipeline and does not establish causal wildfire-to-cascade prediction.
\section{Results}
\subsection{Calibration Audit}
\label{subsec:calibration_audit}
A three-video mechanism audit used $\alpha=0.5$, $\beta=0.25$, and $\gamma=4$. Table~\ref{tab:calibration} shows that the no-fire sequence remains zero. Previous legacy per-video min--max normalization results in both detected fire sources being forced to a maximum of one. Under the shared mapping, their maxima remain source-dependent. This verifies the removal of the forced-unit-maximum artifact, but does not by itself establish physical fire calibration or improved grid prediction. 

\begin{table}[t]
\centering
\caption{Environmental-signal calibration audit}
\label{tab:calibration}
\begin{tabular}{lccc}
\toprule
Source & Legacy max. & Shared mean & Shared max.\\
\midrule
No fire & 0.000 & 0.000 & 0.000\\
Night fire & 1.000 & 0.579 & 0.858\\
Wildfire & 1.000 & 0.625 & 0.814\\
\bottomrule
\end{tabular}
\end{table}

\subsection{Held-Out Source Diagnostic}
The 40-scenario source-disjoint test produced 10 true positives, 8 false positives, 22 true negatives, and no false negatives. As summarized in Table~\ref{tab:diagnostic}, all cascade cases conditioned on an unseen night-fire source were detected, while held-out no-fire and industrial-steam cases were rejected. The eight false positives were concentrated in stressed, non-cascading scenarios conditioned on one unseen growing-fire source.

\begin{table}[t]
\centering
\caption{Source-video-disjoint cascade diagnostic}
\label{tab:diagnostic}
\begin{tabular}{lc@{\qquad}lc}
\toprule
Metric & Value & Metric & Value\\
\midrule
Accuracy & 0.800 & Precision & 0.556\\
Recall & 1.000 & $F_1$ & 0.714\\
Specificity & 0.733 & Balanced acc. & 0.867\\
\bottomrule
\end{tabular}
\end{table}

\section{Conclusion}
In summary, this paper presented EnergyEminence, a physics-grounded digital-twin testbed that combines graph-temporal cascade prediction, nonlinear AC simulation, environmental perception, and synchronized replay. A shared environmental-evidence calibration improved cross-video comparability, while source-video-disjoint evaluation exposed a growing-fire shortcut-learning failure that was hidden by scenario-level random splitting. The result demonstrates that environmental hazard evidence should not be treated as direct cascade ground truth. Instead, environmental alerts and grid-only GNN/LSTM inference should remain independently testable and converge at the digital-twin decision layer. Although evaluated on synthetic videos, one IEEE 118-bus-style topology, and a limited source set, the study provides a reproducible, data-driven evaluation methodology for explainable grid monitoring and motivates modular multimodal architectures for future utility-scale validation.

\bibliographystyle{IEEEtran}
\bibliography{refs}

\end{document}